# Automated Prostate Gland Segmentation in MRI Using nnU-Net

Pablo Rodriguez-Belenguer[1], Gloria Ribas[1], Javier Aquerreta Escribano[1], Rafael Moreno-Calatayud[1], Leonor Cerdá-Alberich[1], Luis Martí-Bonmatí[1]

[1]Biomedical Imaging Research Group (GIBI2[30]), Instituto de Investigación Sanitaria La Fe, Valencia, Spain

# Abstract

Accurate segmentation of the prostate gland in multiparametric MRI (mpMRI) is a fundamental step for a wide range of clinical and research applications, including image registration, volume estimation, and radiomic analysis. However, manual delineation is time-consuming and subject to inter-observer variability, while general-purpose segmentation tools often fail to provide sufficient accuracy for prostate-specific tasks.

In this work, we propose a dedicated deep learning–based approach for automatic prostate gland segmentation using the nnU-Net v2 framework. The model leverages multimodal mpMRI data, including T2-weighted imaging, diffusion-weighted imaging (DWI), and apparent diffusion coefficient (ADC) maps, to exploit complementary tissue information. Training was performed on 981 cases from the PI-CAI dataset using whole-gland annotations, and model performance was assessed through 5-fold cross-validation and external validation on an independent cohort of 54 patients from Hospital La Fe.

The proposed model achieved a mean Dice score of 0.96 ± 0.00 in cross-validation and 0.82 on the external test set, demonstrating strong generalization despite domain shift. In comparison, a general-purpose approach (TotalSegmentator) showed substantially lower performance, with a Dice score of 0.15, primarily due to under-segmentation of the gland.

These results highlight the importance of task-specific, multimodal segmentation strategies and demonstrate the potential of the proposed approach for reliable integration into clinical research workflows. To facilitate reproducibility and deployment, the model has been fully containerized and is available as a ready-to-use inference tool.

# Introduction

Prostate cancer is the most frequently diagnosed malignancy and the second leading cause of cancer-related death among men worldwide [1]. Diagnosis typically begins with elevated prostate-specific antigen (PSA) levels or abnormal digital rectal examination, followed by confirmatory transrectal or transperineal biopsy. Histological grading based on the Gleason score informs the ISUP Grade Groups and helps distinguish clinically significant from insignificant disease [2]. While biopsy remains the gold standard, it is invasive and susceptible to sampling errors. As a result, multiparametric MRI (mpMRI) has become a cornerstone in non-invasive prostate cancer detection, localization, and characterization, with PI-RADS providing a standardized framework to assess the likelihood of clinically significant cancer [3].

Despite these advances, mpMRI-based workflows still rely on accurate anatomical delineation of the prostate gland to enable reliable quantitative analysis. In particular, prostate gland segmentation is a key step for tasks such as image registration, volume estimation, radiomic feature extraction, and region-of-interest definition. Manual delineation, however, is time-consuming and subject to inter-observer variability, which can limit reproducibility and scalability in both clinical and research settings.

In this context, automatic segmentation of the prostate gland is of critical importance for a wide range of downstream tasks in mpMRI analysis. In our research group (GIBI2$^{30}$), these tasks are routinely performed as part of different research pipelines, making accurate and consistent gland delineation a fundamental requirement.

Although general-purpose tools such as TotalSegmentator are readily available, their performance may not be optimal for prostate-specific applications. TotalSegmentator was originally developed as a large-scale, multi-organ segmentation framework trained on heterogeneous MRI and CT datasets to segment a wide range of anatomical structures (up to 80) across varying imaging conditions [4]. While this design promotes broad generalization, it may limit task-specific accuracy for smaller or clinically challenging structures such as the prostate gland. In addition, such approaches typically rely on single-sequence inputs, failing to exploit the complementary information provided by multiparametric MRI. In our experience, this results in segmentations that are not sufficiently accurate or robust for routine use in downstream analysis pipelines.

To overcome these limitations, we propose a dedicated prostate gland segmentation model tailored to our use case. In contrast to general-purpose approaches, our method leverages the complementary information from multiple MRI sequences, including T2-weighted imaging, diffusion-weighted imaging (DWI), and apparent diffusion coefficient (ADC) maps, which provide synergistic information for prostate tissue characterization. The model is trained on the PI-CAI dataset using the nnU-Net V2 framework, with the aim of delivering a reliable, fully automated, and reproducible tool suitable for integration into daily clinical research workflows. Both the proposed model and TotalSegmentator are subsequently evaluated on an independent cohort of 54 patients from Hospital La Fe, enabling a direct and fair comparison under real-world clinical conditions.

# Methods

## Data

### Prostate Imaging – Cancer AI dataset (for training)

The segmentation model was developed using the Public Training and Development Dataset from the Prostate Imaging – Cancer AI (PI-CAI) challenge [5]. This publicly available resource contains anonymized mpMRI studies provided by the challenge, of which 981 were used in this study, each including at least axial T2-weighted (T2W), high b-value diffusion-weighted imaging (DWI ≥1000 s/mm²), and ADC maps. Optional sagittal and coronal T2W acquisitions are also present. MRI scans were acquired using surface coils on Siemens and Philips systems, and basic clinical data are provided, including patient age, prostate volume, PSA level, PSA density, and ISUP grade group. The imaging data are hosted on Zenodo (DOI: [10.5281/zenodo.6624726](10.5281/zenodo.6624726)), and the associated segmentation annotations are available via the official [PI-CAI GitHub](PI-CAI GitHub) repository.

The cohort has a mean age of 65.6 years (range: 35–92), with a mean prostate volume of 64.6 mL (20–155 mL) and a mean PSA level of 12.2 ng/mL (0.1–224 ng/mL). A total of 544 cases were negative (ISUP 0), while 646 were positive for prostate cancer, distributed as follows: ISUP 1 (n=152), ISUP 2 (n=159), ISUP 3 (n=63), ISUP 4 (n=28), and ISUP 5 (n=35).

This model was trained using AI-generated prostate masks available for all 981 selected cases ([Bosma22b whole gland masks](Bosma22b whole gland masks)) [6].

### Hospital La Fe dataset (for testing)

This dataset is the result of a retrospective study conducted within the DETERMIA project (approval no. PMPTA23/00010), approved by the institutional ethics committee, which waived informed consent. Patients were consecutively collected from Hospital La Fe between January 2015 and January 2023. After applying quality-control criteria to ensure the availability of T2-weighted and diffusion-weighted imaging, a total of 1,386 patients were retained.

MRI examinations were acquired using surface coils across multiple vendors (GE, Siemens, and Philips), predominantly at 3T, with a smaller proportion at 1.5T. Diffusion-weighted imaging (DWI) was acquired with b-values of 0 and 1000 s/mm² in 67.5% of patients. In the remaining cases, variable b-value acquisitions were standardized to b = 1000 s/mm² using a mono-exponential signal model (https://github.com/NIH-MIP/PyComputeBValue). ADC maps were subsequently computed from the processed diffusion data, ensuring consistency across the dataset.

From this cohort, a subset of 54 patients with expert-validated prostate gland segmentations was selected for evaluation. All selected cases include multiparametric MRI acquisitions (T2W, DWI, and ADC), making them suitable for direct comparison with the proposed model. This independent dataset is used exclusively for testing and enables a direct comparison between the proposed approach and TotalSegmentator in a real-world clinical setting.

## Preprocessing

For the PI-CAI dataset, minimal preprocessing was applied. All images were spatially resampled to a common reference grid, using the T2-weighted image as the anatomical reference. This ensures consistent voxel spacing and spatial alignment across modalities, enabling voxel-wise correspondence while preserving the original image characteristics.

In contrast, a more extensive preprocessing pipeline was applied to the Hospital La Fe dataset to ensure harmonization across heterogeneous acquisitions. First, denoising was performed using anisotropic diffusion filtering for T2-weighted images with SimpleITK (v2.5.0), and SUSAN filtering for diffusion-weighted imaging using Nipype (v1.10.0) interfacing FSL. Intensity inhomogeneity was corrected using N4 bias field correction.

Subsequently, spatial alignment was carried out using Elastix, applying rigid registration within diffusion-weighted volumes and affine registration between T2-weighted and diffusion-weighted sequences. Finally, ADC maps were computed from the co-registered diffusion images.

## Model segmentation

To enable automatic delineation of the prostate gland in mpMRI studies, a 3D segmentation model was developed using the nnU-Net framework, configured in its 3D full-resolution mode with a residual encoder backbone. This configuration allows the model to fully exploit volumetric context while maintaining high spatial detail.

The model was trained from scratch on 981 cases from the PI-CAI dataset, using whole-gland annotations to learn robust anatomical representations across heterogeneous imaging conditions. Training was conducted for 250 epochs under a cross-validation scheme to enhance generalization. The optimization was performed using a composite loss function combining Dice and Cross-Entropy terms with equal weighting.

The resulting model has an approximate size of 80 GB and is designed to operate in a fully automated manner. For evaluation, performance was assessed on an independent cohort of 54 patients from Hospital La Fe, enabling the analysis of generalization capability in a real-world clinical setting.

## Model validation and evaluation

This model was validated using a consistent strategy combining internal cross-validation and external testing on independent data. A 5-fold cross-validation (CV) was conducted during training using nnU-Net's default configuration. This allowed for robust estimation of model performance across different subsets of the training data.

Validation was performed using the following segmentation metrics:

- Dice Similarity Coefficient (DSC)
- Intersection over Union (IoU)
- Hausdorff Distance (HD)
- Precision

- Sensitivity

For each fold, validation metrics were computed individually and reported to assess the model's internal generalization. Final evaluation was conducted on independent test sets: 54 manually annotated prostate glands and 16 manually annotated lesion cases from Hospital La Fe. These test sets provided an external validation of performance in a real-world clinical setting.

## Implementation and availability

To facilitate reproducibility and practical deployment, the proposed model has been fully containerized using Docker. The tool provides an end-to-end automated pipeline for inference, including model loading, preprocessing compatibility, and segmentation prediction.

The containerized version of the framework is publicly available and can be directly used for inference on new datasets. Detailed instructions for installation, configuration, and usage are provided in the following repository:

https://bitbucket.org/gibi230/prostate_gland_segmentation/src/master/

This implementation enables seamless integration into existing research workflows and allows users to perform prostate gland segmentation without requiring manual setup of dependencies or model configuration.

# Results and discussion

Table 1 summarizes the quantitative performance of the proposed prostate gland segmentation model across the five cross-validation folds on the PI-CAI dataset. The model achieved a mean Dice score of 0.96 ± 0.00 and an IoU of 0.93 ± 0.01, indicating excellent volumetric agreement with reference annotations. The Hausdorff Distance was 4.60 ± 0.75 mm, reflecting accurate boundary delineation, while sensitivity and precision were both 0.96 ± 0.01, demonstrating a well-balanced segmentation performance. These results highlight the robustness and consistency of the model under controlled training conditions.

**Table 1.** Gland segmentation performance in cross-validation (PI-CAI).

| Fold | Dice | IoU | HD | Sensitivity | Precision |
| --- | --- | --- | --- | --- | --- |
| Fold 0 | 0.96 | 0.94 | 4.98 | 0.97 | 0.96 |
| Fold 1 | 0.96 | 0.93 | 3.94 | 0.96 | 0.96 |
| Fold 2 | 0.95 | 0.92 | 5.74 | 0.95 | 0.95 |
| Fold 3 | 0.96 | 0.93 | 4.80 | 0.96 | 0.97 |
| Fold 4 | 0.96 | 0.93 | 3.98 | 0.96 | 0.96 |
| Mean ± STD | 0.96 ± 0.00 | 0.93 ± 0.01 | 4.60 ± 0.75 | 0.96 ± 0.01 | 0.96 ± 0.01 |

A quantitative comparison between the proposed model and TotalSegmentator on the external Hospital La Fe cohort is presented in Table 2. The proposed model achieved a Dice score of 0.82 and an IoU of 0.70, with a Hausdorff Distance of 11.35 mm, while sensitivity and precision reached 0.87 and 0.78, respectively. In contrast, TotalSegmentator showed substantially lower overlap performance, with a Dice score of 0.15 and an IoU of 0.10, along with a higher Hausdorff Distance of 23.83 mm, indicating poorer boundary delineation.

Although TotalSegmentator achieved higher precision (0.91), this was accompanied by a very low sensitivity (0.10), revealing a strong tendency toward under-segmentation. In comparison, the proposed model maintains a more balanced trade-off between sensitivity and precision, resulting in more complete and clinically meaningful gland delineations.

**Table 2.** Results over test set

| Fold | Dice | IoU | HD | Sensitivity | Precision |
| --- | --- | --- | --- | --- | --- |
| Proposed model | 0.82 | 0.70 | 11.35 | 0.87 | 0.78 |
| TotalSegmentator | 0.15 | 0.10 | 23.83 | 0.10 | 0.91 |

Figure 1 illustrates qualitative examples of prostate gland segmentation results for both the proposed model and TotalSegmentator. A subset of five representative patients from the external Hospital La Fe cohort was selected to provide a visual comparison. For each patient, the axial slice was chosen based on the ground truth segmentation, specifically selecting the slice containing the largest number of segmented voxels, ensuring that the most informative anatomical region of the prostate gland is displayed.

Each row corresponds to a different patient, while columns represent, from left to right, the original T2-weighted image, the expert-validated ground truth segmentation, the prediction obtained with the proposed nnU-Net-based model, and the segmentation produced by TotalSegmentator. This layout allows for a direct visual comparison of segmentation quality across methods.

As can be observed, the proposed model provides segmentations that closely match the ground truth, both in terms of shape and spatial extent. In contrast, TotalSegmentator frequently underestimates the prostate gland, producing incomplete segmentations that fail to capture its full anatomical boundaries. These qualitative findings are consistent with the quantitative results reported in Table 2, particularly the lower Dice scores and sensitivity values observed for TotalSegmentator.

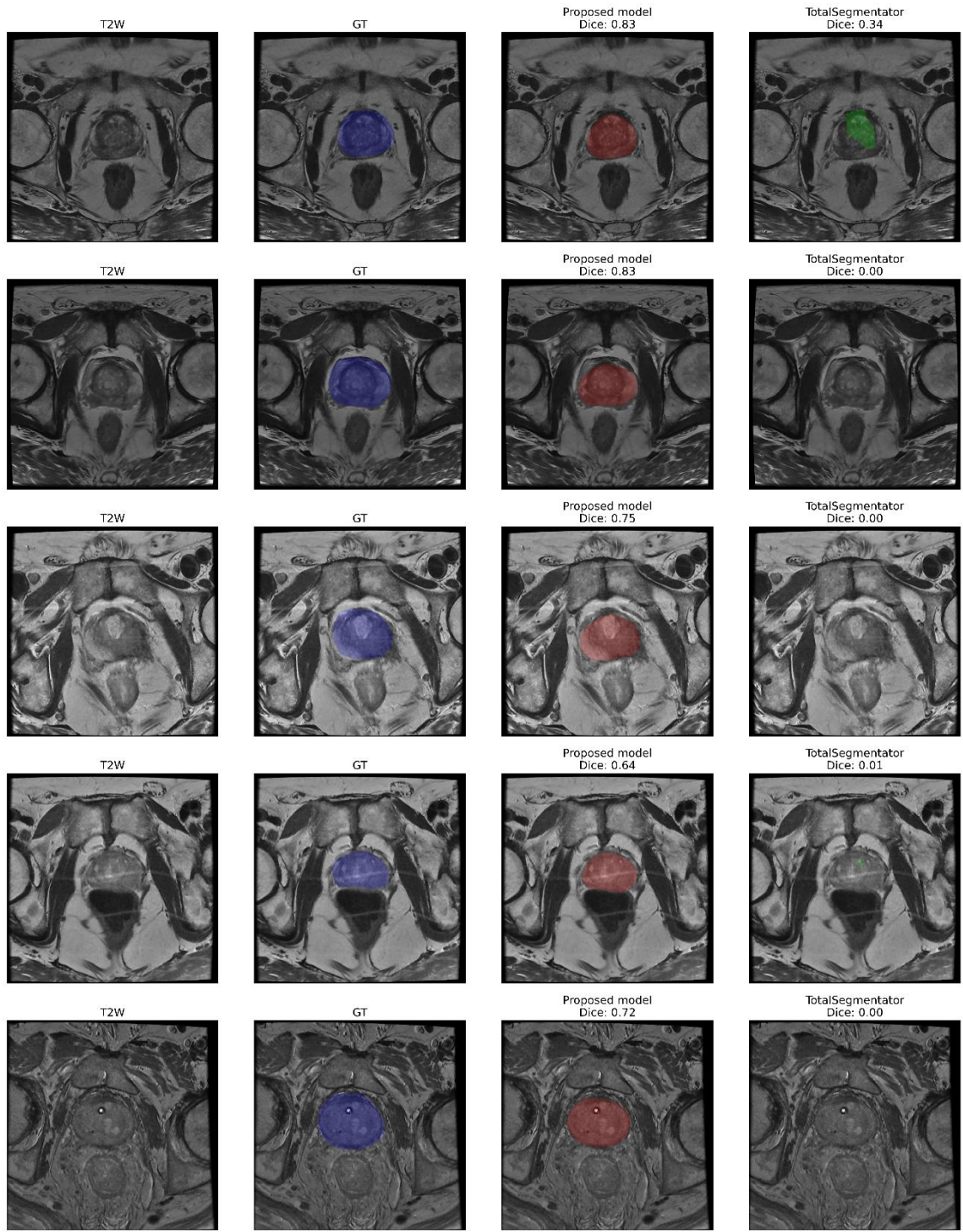

**Figure 1.** Qualitative comparison of prostate gland segmentation on the Hospital La Fe cohort. Each row corresponds to a different patient. From left to right: T2-weighted image, ground truth, proposed model, and TotalSegmentator. The slice was selected based on the ground truth mask (maximum voxel area). Dice scores are reported for the predicted segmentations.

Overall, these results demonstrate that, despite the challenges introduced by domain shift, the proposed model maintains strong generalization capability and significantly outperforms TotalSegmentator in a real-world clinical setting. The improved balance between sensitivity and precision further supports its suitability for downstream tasks requiring accurate and complete gland delineation.

# Conclusions

In this work, we presented a dedicated deep learning–based framework for automatic prostate gland segmentation in multiparametric MRI using the nnU-Net architecture. The proposed approach leverages multimodal information, including T2-weighted imaging, diffusion-weighted imaging, and ADC maps, to provide accurate and consistent gland delineations.

The model achieved excellent performance in cross-validation on the PI-CAI dataset, demonstrating high accuracy and robustness under controlled conditions. When evaluated on an independent external cohort from Hospital La Fe, a decrease in performance was observed due to domain shift; however, the model maintained strong generalization capability, confirming its applicability in real-world clinical scenarios.

Importantly, the proposed method significantly outperformed a general-purpose segmentation approach such as TotalSegmentator, both quantitatively and qualitatively. While TotalSegmentator exhibited high precision, it consistently failed to capture the full extent of the prostate gland, leading to severe under-segmentation. In contrast, our model achieved a more balanced trade-off between sensitivity and precision, resulting in more complete and clinically meaningful segmentations.

# CRediT authorship contribution statement

**Pablo Rodríguez-Belenguer:** Writing – review & editing, Writing – original draft, Visualization, Validation, Software, Methodology, Formal analysis, Conceptualization. **Gloria Ribas:** Data curation. **Javier Aquerreta Escribano:** Data curation. **Rafael Moreno-Calatayud:** Writing – review & editing. **Leonor Cerdá-Alberich:** Writing – review & editing and supervision. **Luis Martí-Bonmatí:** Writing – review & editing, Conceptualization.

# Ethics statement

This study was conducted in accordance with institutional guidelines and approved by the ethics committee within the DETERMIA project (approval no. PMPTA23/00010), which waived the requirement for informed consent due to the retrospective nature of the study. All data were anonymized prior to analysis.

The training data were obtained from the publicly available PI-CAI dataset, which is fully anonymized and distributed for research purposes.

# Declaration of competing interest

The authors declare that they have no known competing financial interests or personal relationships that could have appeared to influence the work reported in this paper.

## Acknowledgements

This project has received funding from Determia with code ISCII PMPTA23/00010 granted by the Instituto de Salud Carlos III call for Research, Development, and Innovation (R&D&i) Projects related to Personalized Medicine and Advanced Therapies (Transmissions Initiative), co-financed by the European Union-NextGenerationEU /Recovery Plan, transformation and Resilence (RPTR).

## References

1.  Bray F, Laversanne M, Sung H, et al (2024) Global cancer statistics 2022: GLOBOCAN estimates of incidence and mortality worldwide for 36 cancers in 185 countries. CA Cancer J Clin 74:229–263. https://doi.org/10.3322/caac.21834

2.  Cornford P, Bergh RCN van den, Briers E, et al (2024) EAU-EANM-ESTRO-ESUR-ISUP-SIOG Guidelines on Prostate Cancer—2024 Update. Part I: Screening, Diagnosis, and Local Treatment with Curative Intent. Eur Urol 86:148–163. https://doi.org/10.1016/j.eururo.2024.03.027

3.  Weinreb JC, Barentsz JO, Choyke PL, et al (2016) PI-RADS Prostate Imaging – Reporting and Data System: 2015, Version 2. Eur Urol 69:16–40. https://doi.org/10.1016/j.eururo.2015.08.052

4.  Akinci D'Antonoli T, Berger LK, Indrakanti AK, et al (2025) TotalSegmentator MRI: Robust Sequence-independent Segmentation of Multiple Anatomic Structures in MRI. Radiology 314:e241613. https://doi.org/10.1148/radiol.241613

5.  Saha A, Bosma JS, Twilt JJ, et al (2024) Artificial intelligence and radiologists in prostate cancer detection on MRI (PI-CAI): an international, paired, non-inferiority, confirmatory study. Lancet Oncol 25:879–887. https://doi.org/10.1016/S1470-2045(24)00220-1

6.  Bosma JS, Saha A, Hosseinzadeh M, et al (2023) Semisupervised Learning with Report-guided Pseudo Labels for Deep Learning–based Prostate Cancer Detection Using Biparametric MRI. Radiol Artif Intell 5:e230031. https://doi.org/10.1148/ryai.230031

7.  Pooch EHP, Agrotis G, Cai L, et al (2025) Semi-Supervised Learning in Prostate MRI Tumor Segmentation Approaches Fully-Supervised Performance on External Validation. medRxiv. https://doi.org/10.1101/2025.05.13.25327456